\documentclass[10pt,twocolumn,letterpaper]{article}

\usepackage{cvpr}
\usepackage{times}
\usepackage{epsfig}
\usepackage{graphicx}
\usepackage{amsmath}
\usepackage{amssymb}
\usepackage{multirow}
\usepackage{makecell}
\usepackage{subcaption}
\usepackage{mathtools}
\DeclarePairedDelimiter{\ceil}{\lceil}{\rceil}
\DeclarePairedDelimiter\floor{\lfloor}{\rfloor}

\usepackage[pagebackref=true,breaklinks=true,letterpaper=true,colorlinks,bookmarks=false]{hyperref}

 \cvprfinalcopy 

\def\cvprPaperID{2058} 

\ifcvprfinal\fi
\begin{document}

\title{Recurrent Slice Networks for 3D Segmentation of Point Clouds}

\author{Qiangui Huang
 \qquad Weiyue Wang \qquad Ulrich Neumann\\
University of Southern California \\
Los Angeles, California \\
{\tt\small \{qianguih,weiyuewa,uneumann\}@usc.edu} 
}

\maketitle

\begin{abstract}
Point clouds are an efficient data format for 3D data. However, existing 3D segmentation methods for point clouds either do not model local dependencies \cite{pointnet} or require added computations \cite{kd-net,pointnet2}. This work presents a novel 3D segmentation framework, RSNet\footnote{Codes are released here https://github.com/qianguih/RSNet}, to efficiently model local structures in point clouds. The key component of the RSNet is a lightweight local dependency module. It is a combination of a novel slice pooling layer, Recurrent Neural Network (RNN) layers, and a slice unpooling layer. The slice pooling layer is designed to project features of unordered points onto an ordered sequence of feature vectors so that traditional end-to-end learning algorithms (RNNs) can be applied. The performance of RSNet is validated by comprehensive experiments on the S3DIS\cite{stanford}, ScanNet\cite{scannet}, and ShapeNet \cite{shapenet} datasets. In its simplest form, RSNets surpass all previous state-of-the-art methods on these benchmarks. And comparisons against previous state-of-the-art methods \cite{pointnet, pointnet2} demonstrate the efficiency of RSNets.

\end{abstract}


\section{Introduction}

Most 3D data capturing devices (like LiDAR and depth sensors) produce point clouds as raw outputs. However, there are few state-of-the-art 3D segmentation algorithms that use point clouds as inputs. The main obstacle is that point clouds are usually unstructured and unordered, so it is hard to apply powerful end-to-end learning algorithms. As a compromise, many researchers transform point clouds into alternative data formats such as voxels \cite{segcloud, 3d_shapenet, voxnet, charles_vol_mtv, weiyue} and multi-view renderings \cite{charles_vol_mtv, mv-cnn, pang_icpr}.

Unfortunately, information loss and quantitation artifacts often occur in data format transformations. These can lead to 3D segmentation performance drops as a result due to loss of local contexts. Moreover, the 3D CNNs \cite{segcloud, 3d_shapenet, voxnet, charles_vol_mtv, jing} and 2D multi-view CNNs \cite{mv-cnn, charles_vol_mtv} designed for these data formats are often time- and memory- consuming.

\begin{figure}[t]
\begin{center}
   \includegraphics[width=1.0\linewidth]{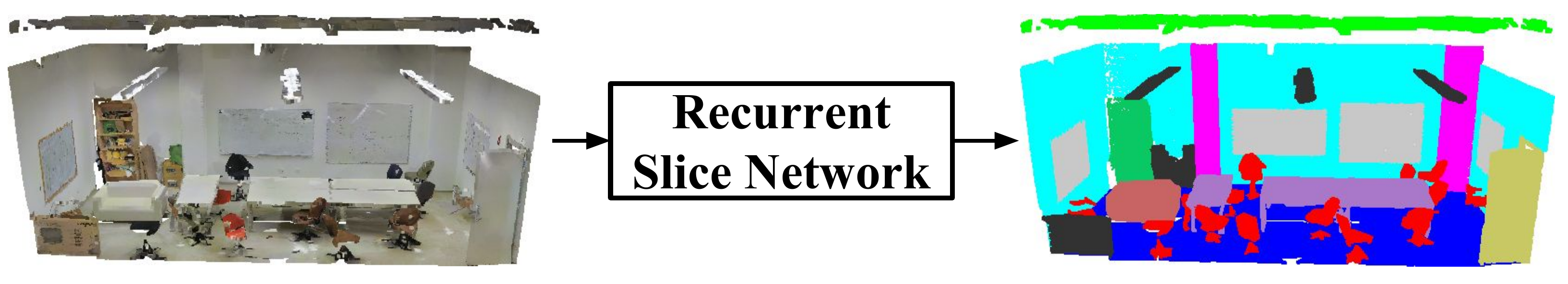}
\end{center}
   \caption{The RSNet takes raw point clouds as inputs and outputs semantic labels for each point.}
\label{fig:example}
\end{figure}

In this paper, we approach 3D semantic segmentation tasks by directly dealing with point clouds. A simple network, a Recurrent Slice Network (RSNet), is designed for 3D segmentation tasks. As shown in Fig.\ref{fig:example}, the RSNet takes as inputs raw point clouds and outputs semantic labels for each of them.

The main challenge in handling point clouds is modeling local geometric dependencies. Since points are processed in an unstructured and unordered manner, powerful 2D segmentation methods like Convolutional Neural Networks (CNN) and Recurrent Neural Networks (RNNs) cannot be directly generalized to them.


In RSNets, the local context problem is solved by first projecting unordered points into ordered features and then applying traditional end-to-end learning algorithms. The projection is achieved by a novel slice pooling layer. In this layer, the inputs are features of unordered points and the output is an ordered sequence of aggregated features. Next, RNNs are applied to model dependencies in this sequence. Finally, a slice unpooling layer assigns features in the sequence back to points. In summary, the combination of the slice pooling layer, RNN layers, and the slice unpooling layer forms the local dependency module in RSNets. We note that the local dependency module is highly efficient. As shown in Section 3.2, the time complexity of the slice pooling/unpooling layer is $O(n)$ w.r.t the number of input points and $O(1)$ w.r.t the local context resolutions.

The performances of RSNets are validated on three challenging benchmarks. Two of them are large-scale real-world datasets, the S3DIS dataset \cite{stanford} and the ScanNet dataset \cite{scannet}. Another one is the ShapeNet dataset \cite{shapenet}, a synthetic dataset. RSNets outperform all prior results and significantly improve performances on the S3DIS and ScanNet datasets.

In following parts of the paper, we first review related works in Section 2. Then, details about the RSNet are presented in Section 3. Section 4 reports all experimental results and Section 5 draws conclusions.

\section{Related Works}
Traditional 3D analysis algorithms are based on hand-crafted features \cite{pang_icpr, pang_3dv, pang_mva, pang_3dv2, pand_wacv2, qiu_eccv, qiu_wacv, qiu_3dv, huang_cvpr, huang_3dv, huang_icra}. Recently, there are some works that utilize end-to-end learning algorithms for 3D data analysis. They are categorized by their input data formats as follows.

\textbf{Voxelized Volumes}. \cite{3d_shapenet, voxnet, charles_vol_mtv, huang2016point, jing} made the early attempts of applying end-to-end deep learning algorithms for 3D data analysis, including 3D shape recognition, 3D urban scene segmentation \cite{jing}. They converted raw point cloud data into voxelized occupancy grids and then applied 3D deep Convolutional Neural Networks to them. Due to the memory constraints of 3D convolutions, the size of input cubes in these methods was limited to $60^3$ and the depth of the CNNs was relatively shallow. Many works have been proposed to ease the computational intensities. One direction is to exploit the sparsity in voxel grids. In \cite{vote3deep}, the authors proposed to calculate convolutions at sparse input locations by pushing values to their target locations. Benjamin Graham designed a sparse convolution network \cite{scnn1, scnn2} and applied it for 3D segmentation tasks \cite{yi_shapnet}. \cite{fpnn} tried to reduce computation by sampling 3D data at sparse points before feeding them into networks. In \cite{OctNet}, the authors designed a memory efficient data structure, hybrid grid-octree, and corresponding convolution/pooling/unpooling operations to handle higher resolution 3D voxel grids (up to $256^3$). In \cite{segcloud}, the authors managed to consume 3D voxel inputs of higher resolution ($100^3$) and build deeper networks by adopting early down-sampling and efficient convolutional blocks like residual modules. While most of these works were focusing on reducing computational requirements of 3D voxel inputs, few of them tried to deal with the quantitation artifacts and information loss in voxelization.

\textbf{Multi-view Renderings}. Another popular data representation for 3D data is its multi-view rendering images. \cite{pang_icpr} designed a multi-view CNN for object detection in point clouds. In \cite{deeppano}, 3D shapes were transformed into panoramic views, i.e., a cylinder project around its principal axis. \cite{mv-cnn} designed a 2D CNN for 3D shape recognition by taking as inputs multi-view images. In \cite{charles_vol_mtv}, the authors conducted comprehensive experiments to compare the recognition performances of 2D multi-view CNNs against 3D volumetric CNNs. More recently, multi-view 2D CNNs have been applied to 3D shape segmentation and achieved promising results. Compared to volumetric methods, multi-view based methods require less computational costs. However, there is also information loss in the multi-view rendering process.

\textbf{Point Clouds}. In the seminal work of PointNet \cite{pointnet}, the authors designed a network to consume unordered and unstructured point clouds. The key idea is to process points independently and then aggregate them into a global feature representation by max-pooling. PointNet achieved state-of-the-art results on several 3D classification and segmentation tasks. However, there were no local geometric contexts in PointNet. In the following work, PointNet++ \cite{pointnet2}, the authors improved PointNet by incorporating local dependencies and hierarchical feature learning in the network. It was achieved by applying iterative farthest point sampling and ball query to group input points. In another direction, \cite{kd-net} proposed a KD-network for 3D point clouds recognition. In KD-network, a KD-tree was first built on input point clouds. Then, hierarchical groupings were applied to model local dependencies in points.

Both works showed promising improvements on 3D classification and segmentation tasks, which proved the importance of local contexts. However, their local context modeling methods all relied on heavy extra computations such as the iterate farthest point sampling and ball query in \cite{pointnet2} and the KD-tree construction in \cite{kd-net}. More importantly, their computations will grow linearly when higher resolutions of local details are used. For example, higher local context resolutions will increase the number of clusters in \cite{pointnet2} and result in more computations in iterative farthest point sampling. And higher resolutions will enlarge the kd-tree in \cite{kd-net} which also costs extra computations. In contrast, the key part of our local dependency module, the slice pooling layer, has a time complexity of $O(1)$ w.r.t the local context resolution as shown in Section 3.2.

\section{Method}

Given a set of unordered point clouds $X=\{  x_1, x_2, ..., x_i, ..., x_n  \}$ with $x_i \in \mathbb{R}^d$ and a candidate label set $L=\{  l_1, l_2, ..., l_K \}$, our task is to assign each of input points $x_i$ with one of the $K$ semantic labels. In RSNets, the input is raw point clouds $X$ and output is $Y=\{ y_1, y_2, ..., y_i, ..., y_n  \}$ where $y_i \in L$ is the label assigned to $x_i$. 

\begin{figure}[t]
\begin{center}
   \includegraphics[width=1.0\linewidth]{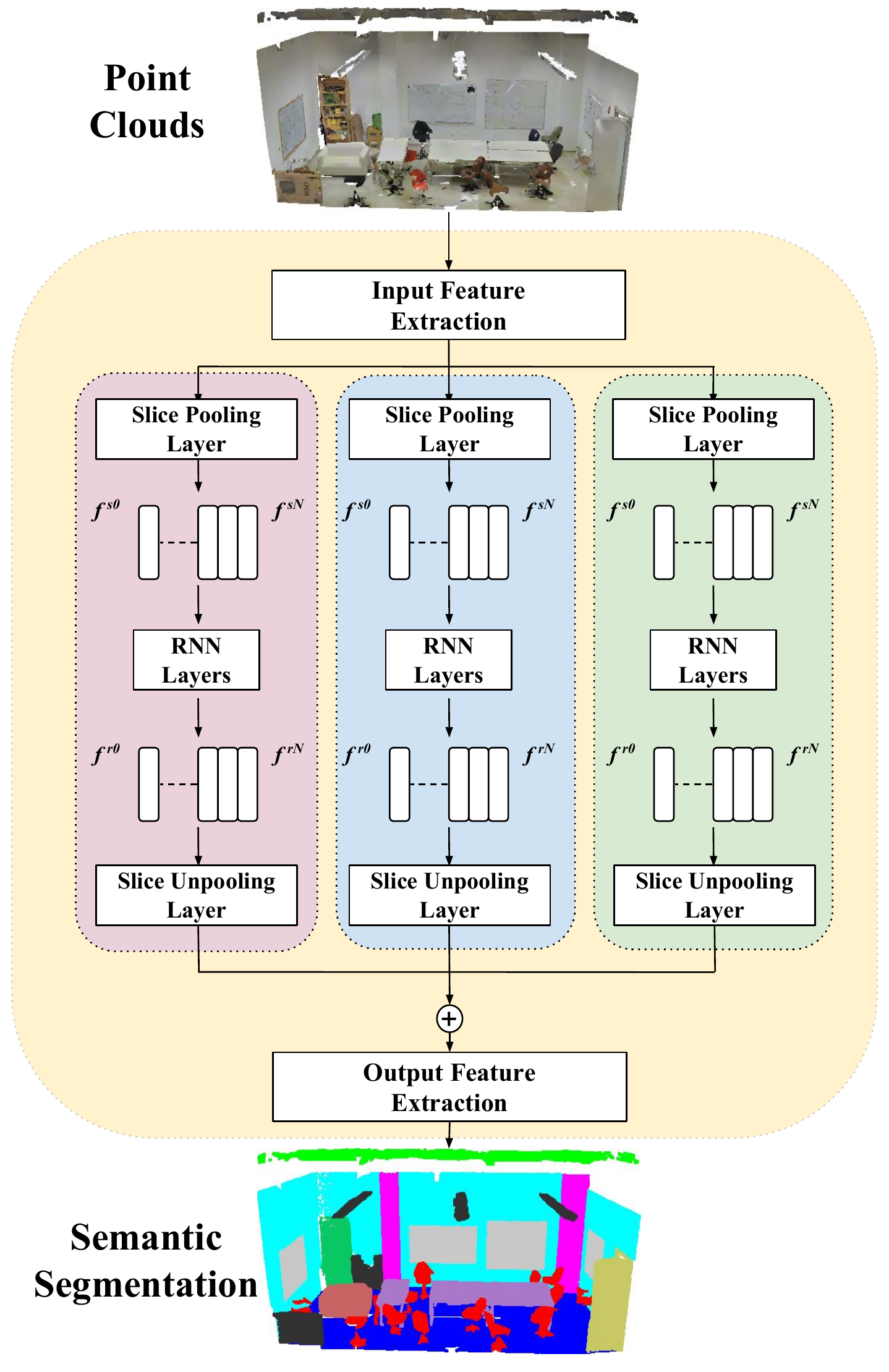}
\end{center}
   \caption{Diagram of an RSNet. The three parallel branches denote the slicing direction along $x$, $y$, and $z$ axis.}
\label{fig:overview}
\end{figure}

A diagram of our method is presented in Fig.\ref{fig:overview}. The input and output feature extraction blocks are used for independent feature generation. In the middle is the local dependency module. Details are illustrated below.



\subsection{Independent Feature Extraction}
There are two independent feature extraction blocks in an RSNet. The input feature block consumes input points $X \in \mathbb{R}^{n \times d^{in}}$ and produce features $F^{in} \in \mathbb{R}^{n \times d^{in}}$. Output feature blocks take processed features $F^{su} \in \mathbb{R}^{n \times d^{su}}$ as inputs and produce final predictions for each point. The superscript \emph{in} and \emph{su} indicate the features are from the input feature block and the slice unpooling layer, respectively. Both blocks use a sequence of multiple $1 \times 1$ convolution layers to produce independent feature representations for each point.

\subsection{Local Dependency Module}

The key part of an RSNet is the local dependency module which is a combination of a slice pooling layer, RNN layers, and a slice unpooling layer. It supports efficient and effective local context modeling. The slice pooling layer is designed to project features of unordered points onto an ordered sequence. RNNs are then applied to model dependencies in the sequence. In the end, the slice unpooling layer reverses the projection and assigns updated features back to each point.

\textbf{Slice Pooling Layer}. The inputs of a slice pooling layer are features of \emph{unordered} point clouds $F^{in} = \{ f^{in}_1,  f^{in}_2, ..., f^{in}_i, ..., f^{in}_n \}$ and the output is an \emph{ordered} sequence of feature vectors. This is achieved by first grouping points into slices and then generating a global representation for each slice via aggregating features of points within the slice.

\begin{figure}[t]
\begin{center}
         \centering
    \begin{subfigure}[t]{0.45\textwidth}
        \centering
     \includegraphics[width=0.9\linewidth]{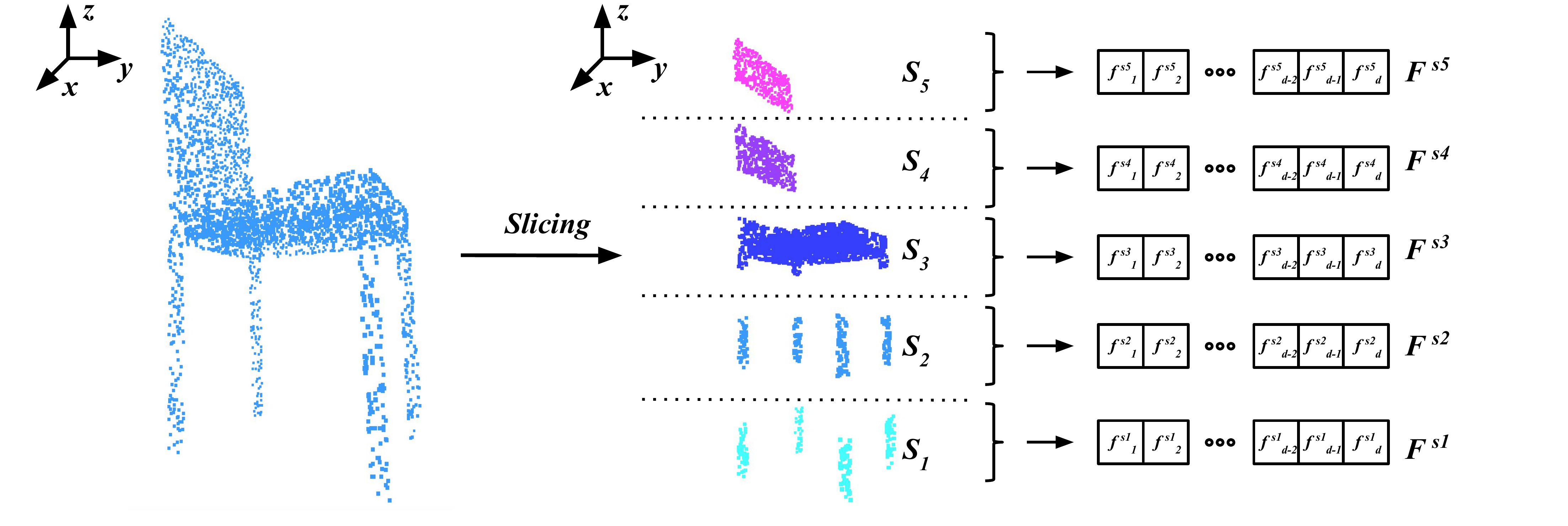}
        \caption{Illustration of the slice pooling operation. A set of points from a chair is used for illustration purpose here.}
    \end{subfigure}%
    
    \begin{subfigure}[t]{0.45\textwidth}
        \centering
     \includegraphics[width=0.9\linewidth]{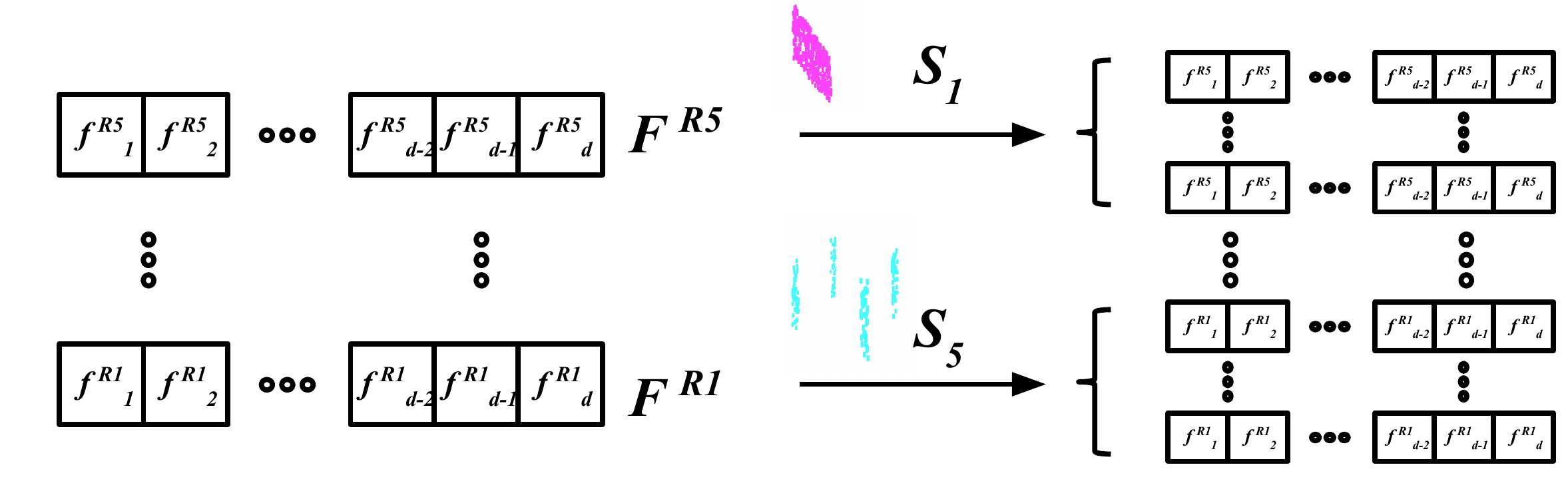}
        \caption{Illustration of the slice unpooling operation. Global feature representation for one point set is replicated back to all points in the set.}
    \end{subfigure}%
    
\end{center}
   \caption{Illustration of slice pooling and slice unpooling operation and RNN modeling for slices.}
\label{fig:slice_pool}
\end{figure}

Three slicing directions, namely slicing along $x$, $y$, and $z$ axis, are considered in RSNets. We illustrate the details of slice pooling operation by taking $z$ axis for example. A diagram of the slice pooling layer is presented in Fig.\ref{fig:slice_pool}. In a slice pooling layer, input points $X$ are first split into slices by their spatial coordinates in $z$ axis. The resolution of each slice is controlled by a hyper-parameter $r$. Assume input points span in the range $[z_{min}, z_{max}]$ in $z$ axis. Then, the point $x_i$ is assigned to the $k^{th}$ slice, where $k = \floor{ (z_i - z_{min} ) / r }$ and $z_i$ is $x_i$'s coordinate in $z$ axis. In total, there are $N$ slices where  $N = \ceil{(z_{max} - z_{min}) / r }$. Here $\ceil{\:}$ and $\floor{\:}$ indicate the ceil and floor function. After this process, all input points are grouped into $N$ slices. They are also treated as $N$ sets of points $S=\{ S_1, S_2, ..., S_i, ..., S_N  \}$, where $S_i$ denotes the set of points assigned to $i^{th}$ slice. In each slice, features of points are aggregated into one feature vector to represent the global information about this slice. Formally, after aggregation, a slice pooling layer produces an ordered sequence of feature vectors $F^s=\{ f^{s1}, f^{s2}, ..., f^{si}, ..., f^{sN}  \}$, where $f^{si}$ is the global feature vector of slice set $S_i$. The max-pooling operation is adopted as the aggregation operator in RSNets. It is formally defined in equation (1).

\begin{gather}
   f^{si} = \underset{ x_j \in S_i }{  \max  }  \{ f^{in}_j \}
\end{gather}

The slice pooling layer has several important properties:
\begin{enumerate}
\item \textbf{Order and Structure}. $F^s$ is an \emph{ordered} and \emph{structured} sequence of feature vectors. In the aforementioned case, $F^s$ is ordered in the $z$ axis. $f^{s1}$ and $f^{sN}$ denote the feature representations of the bottom-most and top-most set of points, respectively. Meanwhile, $f^{si}$ and $f^{s(i-1)}$ are features representing adjacent neighbors. This property makes traditional local dependency modeling algorithms applicable as $F^s$ is structured and ordered now.

\item \textbf{Efficiency}. The time complexity of the slice pooling layer is $O(n)$ ($n$ is the number of the input points). And it is $O(1)$ w. r. t the slicing resolution $r$.

\item \textbf{Local context trade-off}. Given a fixed input, smaller $r$ will produce more slices with richer local contexts preserved while larger $r$ produces fewer slices with coarse local contexts;

\end{enumerate}

\textbf{RNN Layer}. As mentioned above, the slice pooling layer is essentially projecting features of unordered and unstructured input points onto an ordered and structured sequence of feature vectors. RNNs are a group of end-to-end learning algorithms naturally designed for a structured sequence. Thus, they are adopted to model dependencies in the sequence. By modeling one slice as one timestamp, the information from one slice interacts with other slices as the information is flowing through timestamps in RNN units. This enables contexts in slices impact with each other which in turn models the dependencies in them.

In an RSNet, the input of RNN layers is $F^{s}$. In order to guarantee information from one slice could impact on all other slices, RSNets utilize the bidirectional RNN units \cite{brnn} to help information flow in both directions. After processing the inputs with a stack of bidirectional RNNs, the final outputs are $F^r = \{ f^{r1}, f^{r2}, ..., f^{ri}, ..., f^{rN}  \}$ with superscript $r$ denoting the features are from RNN layers. Compared with $F^{s}$, $F^{r}$ has been updated by interacting with neighboring points.

\textbf{Slice Unpooling Layer}.
As the last part of an RSNet's local dependency module, the slice unpooling layer takes updated features $F^r$ as inputs and assigns them back to each point by reversing the projection. This can be easily achieved by storing the slice sets $S$. A diagram of the slice unpooling layer is presented in Fig.\ref{fig:slice_pool}. We note that the time complexity of slice unpooling layer is $O(n)$ w. r. t the number of input points and is $O(1)$ w. r. t slicing resolution as well.

\begin{table*}
\begin{center}
\setlength\tabcolsep{2.0pt}
\begin{tabular}{c|c|c|cccccccccccccc}
\Xhline{3\arrayrulewidth}

Method & mIOU & mAcc & ceiling  & floor & wall & beam & column & window & door & chair & table & bookcase & sofa & board & clutter    \\
\Xhline{3\arrayrulewidth}
PointNet$^A$ \cite{pointnet} & 41.09 & 48.98 & 88.80 & 97.33 & 69.80 & \textbf{0.05} & 3.92 & \textbf{46.26}  & 10.76  & 52,61  & 58.93  & 40.28  & 5.85    & 26.38  & 33.22   \\
3D-CNN \cite{segcloud} & 43.67 & - & - & - & - & - & - & -  & -  & -  & -  & -  & -    & -  & -\\
3D-CNN$^A$ \cite{segcloud} & 47.46 & 54.91 & 90.17 & 96.48 & 70.16 & 0.00 & 11.40 & 33.36  & 21.12  & \textbf{76.12}  & 70.07  & 57.89  & 37.46  & 11.16  & 41.61 \\
3D-CNN$^{AC}$ \cite{segcloud} & 48.92 & 57.35 & 90.06  & 96.05 & 69.86 & 0.00 & \textbf{18.37} & 38.35  & 23.12  & 75.89  & \textbf{70.40}  & \textbf{58.42}  & 40.88  & 12.96  & 41.60\\
Ours & \textbf{51.93} & \textbf{59.42} & \textbf{93.34} & \textbf{98.36} & \textbf{79.18} & 0.00 & 15.75 & 45.37  & \textbf{50.10}  & 65.52  & 67.87  & 22.45  & \textbf{52.45}  & \textbf{41.02}  & \textbf{43.64}\\
\end{tabular}
.
\end{center}
\caption{Results on the Large-Scale 3D Indoor Spaces Dataset (S3DIS). Superscripts $A$ and $C$ denote data augmentation and post-processing (CRF) are used.}
\label{table:s3dis}
\end{table*}

\section{Experiments}
In order to evaluate the performance of RSNets and compare with state-of-the-art, we benchmark RSNets on three datasets, the Stanford 3D dataset (S3DIS) \cite{stanford}, ScanNet dataset \cite{scannet}, and the ShapeNet dataset \cite{shapenet}. The first two are large-scale realistic 3D segmentation datasets and the last one is a synthetic 3D part segmentation dataset. 

We use the strategies in \cite{pointnet, pointnet2} to process all datasets. For the S3DIS and ScanNet datasets, the scenes are first divided into smaller cubes using a sliding window of a fixed size. A fixed number of points are sampled as inputs from the cubes. In this paper, the number of points is fixed as 4096 for both datasets. Then RSNets are applied to segment objects in the cubes. Note that we only divide the scene on the $xy$ plane as in \cite{pointnet}. During testing, the scene is similarly split into cubes. We first run RSNets to get point-wise predictions for each cube, then merge predictions of cubes in the same scene. Majority voting is adopted when multiple predictions of one point are present.

We use one unified RSNet architecture for all datasets. In the input feature extraction block, there are three $1 \times 1$ convolutional layers with output channel number of 64, 64, and 64, respectively. In the output feature extraction block, there are also three $1 \times 1$ convolutional layers with output channel number of 512, 256, and $K$, respectively. Here $K$ is the number of semantic categories. In each branch of the local dependency module, the first layer is a slice pooling layer and the last layer is a slice unpooling layer. The slicing resolution $r$ varies for different datasets. There is a comprehensive performance comparison of different $r$ values in Section 4.2. In the middle are the RNN layers. A stack of 6 bidirectional RNN layers is used in each branch. The numbers of channels for RNN layers are 256, 128, 64, 64, 128, and 256. In the baseline RSNet, Gated Recurrent Unit (GRU) \cite{GRU} units are used in all RNNs.


Two widely used metrics, mean intersection over union (mIOU) and mean accuracy (mAcc), are used to measure the segmentation performances. We first report the performance of a baseline RSNet on the S3DIS dataset. Then, comprehensive studies are conducted to validate various architecture choices in the baseline. In the end, we show state-of-the-art results on the ScanNet and ShapeNet dataset. Through experiments, the performances of RSNets are compared with various state-of-the-art 3D segmentation methods including 3D volumes based methods \cite{segcloud}, spectral CNN based method \cite{spec}, and point clouds based methods \cite{pointnet, pointnet2, kd-net}.

 \begin{figure}[t]
\begin{center}
   \includegraphics[width=1.0\linewidth]{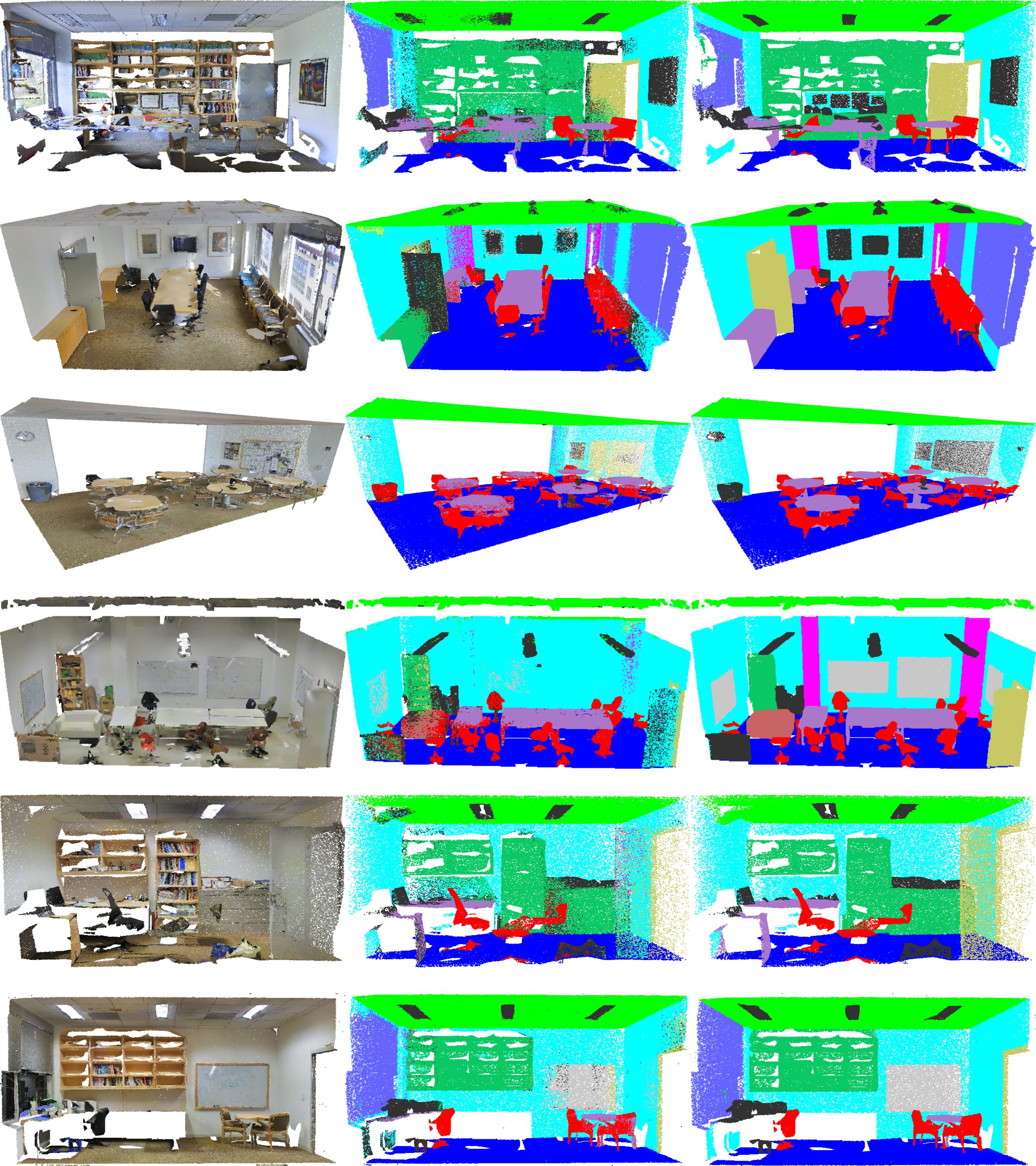}
\end{center}
   \caption{Sample segmentation results on the S3DIS dataset. From left to right are the input scenes, results produced by the RSNet, and ground truth. Best viewed with zoom in.}
\label{fig:stanford}
\end{figure}

\subsection{Segmentation on the S3DIS Dataset}
We first present the performances of a baseline RSNet on the S3DIS dataset. The training/testing split in \cite{segcloud} is used here to better measure the generalization ability of all methods. The slicing resolutions $r$ along the $x$, $y$, $z$ axis are all set at $2cm$. And the block size in $x$ and $y$ axis of each cube is $1m \times 1m$. Given these settings, there are 50 slices ($N=50$) in $x$ and $y$ branch in an RSNet after the slice pooling layer. As we do not limit the block size in $z$ axis, the number of slices along $z$ axis varies on different inputs. In the S3DIS dataset, most of the scenes have a maximum $z$ coordinate around $3m$ which produces around 150 slices.

During testing, the sliding stride is set at $1m$ to generate non-overlapping cubes. The performance of our baseline network is reported in Table.\ref{table:s3dis}. Besides the overall mean IOU and mean accuracy, the IOU of each category is also presented. Meanwhile, some segmentation results are visualized in Fig.\ref{fig:stanford}.

Previous state-of-the-art results \cite{pointnet, segcloud} are reported in Table.\ref{table:s3dis} as well. In \cite{segcloud}, the data representation is voxelized 3D volumes and a 3D CNN is built for segmenting objects in the volumes. Several geometric data augmentation strategies and end-to-end Conditional Random Filed (CRF) are utilized in their work. The PointNet \cite{pointnet} takes the same inputs, point clouds, as our method. It adopted rotation along $z$ axis to augment data. In contrast, our baseline RSN does not use any data augmentations.

The results in Table.\ref{table:s3dis} show that the RSNet has achieved state-of-the-art performances on the S3DIS dataset even without using any data augmentation. In particular, it improves previous 3D volumes based methods \cite{segcloud} by 3.01 in mean IOU and 2.07 in mean accuracy. Compared with prior point clouds based method \cite{pointnet}, it improves the mean IOU by 10.84 and mean accuracy by 10.44. The detailed per-category IOU results show that the RSNet is able to achieve better performances in more than half of all categories (7 out of 13).

We argue that the great performance improvements come from the local dependency module in the RSNet. While PointNet only relies on global features, the RSNet is equipped with local geometric dependencies among points. In summary, the significant performance gains against PointNet demonstrate: 1). local dependency modeling is crucial for 3D segmentation; 2). the combination of the novel slice pooling/unpooling layers and RNN layers is able to support effective spatial dependencies modeling in point clouds. Moreover, the performance improvements against 3D volumes based methods prove that directly handling point clouds can boost the 3D segmentation performances a lot as there are no quantitation artifacts and no local details lost anymore.

\begin{table}
\begin{center}
\begin{tabular}{ccc|c|c}
\Xhline{3\arrayrulewidth}

$r_x$ (cm) & $r_y$ (cm) & $r_z$ (cm)  & mIOU & mAcc    \\
\Xhline{3\arrayrulewidth}
2 & 2 & 1 & 49.12 & 56.63\\
2 & 2 & 2 & \textbf{51.93} & \textbf{59.42}\\
2 & 2 & 5 & 51.20 & 58.97 \\
2 & 2 & 8 & 49.16 & 56.91\\
\hline
1 & 1 & 2 & 49.23 &56.90 \\
2 & 2 & 2 & \textbf{51.93} & \textbf{59.42} \\
4 & 4 & 2 & 48.97 & 57.10\\
6 & 6 & 2 & 47.86 & 56.82\\

\end{tabular}
.
\end{center}
\caption{Varying slice resolutions for RSNs on the S3DIS dataset. $r_x$, $r_y$, and $r_z$ indicate the slicing resolution along $x$, $y$, and $z$ axis, respectively.}
\label{table:resolution}
\end{table}

\begin{table}
\begin{center}
\setlength\tabcolsep{2.0pt}
\begin{tabular}{c|ccc|c|c}
\Xhline{3\arrayrulewidth}

$bs$ (m) & $r_x$ (cm) & $r_y$ (cm) & $r_z$ (cm)  & mIOU & mAcc    \\
\Xhline{3\arrayrulewidth}
\multirow{4}{*}{1} & 2 & 2 & 2 & \textbf{51.93} & \textbf{59.42} \\
& 4 & 4 & 2 & 48.97 & 57.10 \\
& 6 & 6 & 2 & 47.86  & 56.82  \\

\hline
\multirow{4}{*}{2} & 2 & 2 & 2 & 44.15 & 52.39 \\
& 4 & 4 & 2 & \textbf{44.59} & 52.62\\
& 6 & 6 & 2 &43.15  & \textbf{53.07} \\
\hline
\multirow{4}{*}{3} & 2 & 2 & 2 & \textbf{39.08} & \textbf{49.61} \\
& 4 & 4 & 2 & 37.77 & 47.89\\
& 6 & 6 & 2 & 37.55 & 49.01  \\
& 8 & 8 & 2 & 37.21  & 46.35  \\
& 16 & 16 & 2 & 35.25 & 44.70 \\

\end{tabular}
.
\end{center}
\caption{Varying sizes of sliding blocks for RSNs on the S3DIS dataset. $bs$ indicates the block size.}
\label{table:block_size}
\end{table}

\begin{table}
\begin{center}
\setlength\tabcolsep{2.0pt}
\begin{tabular}{c|c|c}
\Xhline{3\arrayrulewidth}

sliding stride during testing & mIOU & mAcc    \\
\Xhline{3\arrayrulewidth}
0.2 & 52.39 & 60.52 \\
0.5 &  \textbf{53.83} & \textbf{61.81} \\
1.0 &  51.93 & 59.42  \\
\end{tabular}
.
\end{center}
\caption{Varying the testing stride on the S3DIS dataset}
\label{table:stride}
\end{table}

\begin{table}
\begin{center}
\setlength\tabcolsep{2.0pt}
\begin{tabular}{c|c|c}
\Xhline{3\arrayrulewidth}

RNN unit & mIOU & mAcc    \\
\Xhline{3\arrayrulewidth}
vanilla RNN &  45.84& 54.82\\
GRU &  \textbf{51.93} & \textbf{59.42} \\
LSTM &  50.08 & 57.80  \\

\end{tabular}
.
\end{center}
\caption{Varying RNN units for RSNs on the S3DIS dataset}
\label{table:rnn}
\end{table}

\begin{table*}
\begin{center}
\setlength\tabcolsep{2.0pt}

\begin{tabular}{cccccccccccc}
\Xhline{3\arrayrulewidth}

Method & mIOU & mAcc & wall  & floor & chair & table & desk & bed & \begin{tabular}{@{}c@{}}book- \\ shelf\end{tabular}   & sofa & sink\\
\Xhline{3\arrayrulewidth}
PointNet \cite{pointnet} 		& 14.69	& 19.90	& 69.44 &88.59 & 35.93&32.78 &2.63 &17.96 & 3.18&32.79 &0.00  \\
PointNet++ \cite{pointnet2}	& 34.26	& 43.77	&77.48& 92.50	& 64.55	& 46.60	& 12.69	&51.32  & 52.93 & 52.27	&30.23\\
\hline
Ours				& \textbf{39.35}	& \textbf{48.37}	&\textbf{79.23}	& \textbf{94.10}	& \textbf{64.99} 	&\textbf{51.04} 	& \textbf{34.53}	&\textbf{55.95} &\textbf{53.02} &\textbf{55.41}& \textbf{34.84} \\

\Xhline{3\arrayrulewidth}
Method  & bathtub & toilet & curtain & counter & door & window & \begin{tabular}{@{}c@{}}shower \\ curtain\end{tabular}  & \begin{tabular}{@{}c@{}}refrid- \\ gerator\end{tabular}  & picture & cabinet & \begin{tabular}{@{}c@{}}other \\ furniture\end{tabular}   \\
\Xhline{3\arrayrulewidth}
PointNet \cite{pointnet} 		& 0.17	&0.00 	&0.00	& 	5.09& 0.00	& 0.00	& 0.00	& 0.00&0.00&4.99&0.13 \\
PointNet++ \cite{pointnet2}	& 42.72	& 31.37	& \textbf{32.97}	& 20.04	& 2.02	&	3.56&27.43 & 18.51 & 0.00 & 23.81 & 2.20 \\
\hline
Ours						& \textbf{49.38}	& \textbf{54.16}	&6.78	& \textbf{22.72}	& \textbf{3.00}	& \textbf{8.75}	&\textbf{29.92} 	& \textbf{37.90}	&\textbf{0.95}	&\textbf{31.29} &\textbf{18.98}\\
\hline
\end{tabular}

\end{center}
\caption{Results on the ScanNet dataset. IOU of each category is also reported here.}
\label{table:scannet}
\end{table*}

\subsection{Ablation Studies}
In this section, we validate the effects of various architecture choices and testing schemes. In particular, several key parameters are considered: 1). the slicing resolution $r$ in RSNets; 2). the size of the sliding block; 3). the sliding stride during testing; 4). the type of RNN units. All settings remain unchanged as the baseline RSNet in following control experiments except explicitly specified.

\textbf{Slicing resolution}. The slicing resolution $r$ is an important hyper-parameter in RSNets. It controls the resolution of each slice which in turn controls how much local details are kept after slice pooling. By using a small slicing resolution, there are more local details preserved as the feature aggregation operation is executed in small local regions. However, a small slicing resolution will produce a large number of slices which requires RNN layers to consume a longer sequence. This may hurt the performance of RSNets as the RNN units may fail to model dependencies in the long sequence due to the ``gradient vanishing" problem \cite{gradient}. On the other hand, a large slicing resolution will eliminate a lot of local details in input data as the feature aggregation is conducted on a wide spatial range. Thus, there is a trade-off of selecting the slicing resolution $r$.

Several experiments are conducted to show the impacts of different slicing resolutions. Two groups of slicing resolutions are tested. In the first group, we fix the slicing resolutions along $x$ and $y$ axis to be $2cm$ and vary the resolution along the $z$ axis. In the second group, the slicing resolution along $z$ axis is fixed as $2cm$ while varying resolutions along $x$ and $y$ axis. Detailed performances are reported in Table.\ref{table:resolution}. Results in Table.\ref{table:resolution} show that the slicing resolution of $2cm$, $2cm$, $2cm$ along $x$, $y$, $z$ axis works best for the S3DIS dataset. Both larger or smaller resolutions decrease the final performances.

\textbf{Size of sliding block}. The size of the sliding block is another key factor in training and testing. Small block sizes may result in too limited contexts in one cube. Large block sizes may put RSNets in a challenging trade-off between slicing resolutions as large block size will either produce more slices when the slicing resolution is fixed or increase the slicing resolution. In Table.\ref{table:block_size}, we report the results of three different block sizes, $1m$, $2m$, and $3m$, along with different slicing resolution choices. The results show that larger block sizes actually decrease the performance. That is because larger block sizes produce a longer sequence of slices for RNN layers, which is hard to model using RNNs. Among various settings, the optimal block size for the S3DIS dataset is $1m$ on both $x$ and $y$ axis.

\textbf{Stride of sliding during testing}. When breaking down the scenes during testing, there are two options, splitting it into non-overlapping cubes or overlapping cubes. In PonintNet \cite{pointnet} , non-overlapping splitting is used while PointNet++ \cite{pointnet2} adopted overlapping splitting. For RSNets, both options are tested. Specifically, we set the sliding stride into three values, $0.2m$, $0.5m$, and $1m$. The first two produce overlapping cubes and the last one produces non-overlapping cubes. All results are reported in Table.\ref{table:stride}. Experimental results show that using overlapped division can slightly increase the performance (0.4$\scriptsize{\sim}$1.9 in mean IOU and 1.1$\scriptsize{\sim}$2.4 in mean accuracy on the S3DIS dataset). However, testing using overlapped division requires more computations as there are more cubes to process. Thus, we select the non-overlap sliding in our baseline RSNet.

\textbf{RNN units}. Due to the ``gradient vanishing" problem in the vanilla RNN unit, two RNN variants, LSTM and GRU, are proposed to model long-range dependencies in inputs. The effects of different RNN units are compared in Table.\ref{table:rnn}. They show that GRU has the best performance for RSNets.

\subsection{Segmentation on the ScanNet dataset}

We now show the performances of RSNets on the ScanNet dataset. The exact same RSNet as Section 4.1 is used to process the ScanNet dataset. The performance of the RSNet is reported in Table.\ref{table:scannet}.

\begin{figure*}[t]
\begin{center}
   \includegraphics[width=1.0\linewidth]{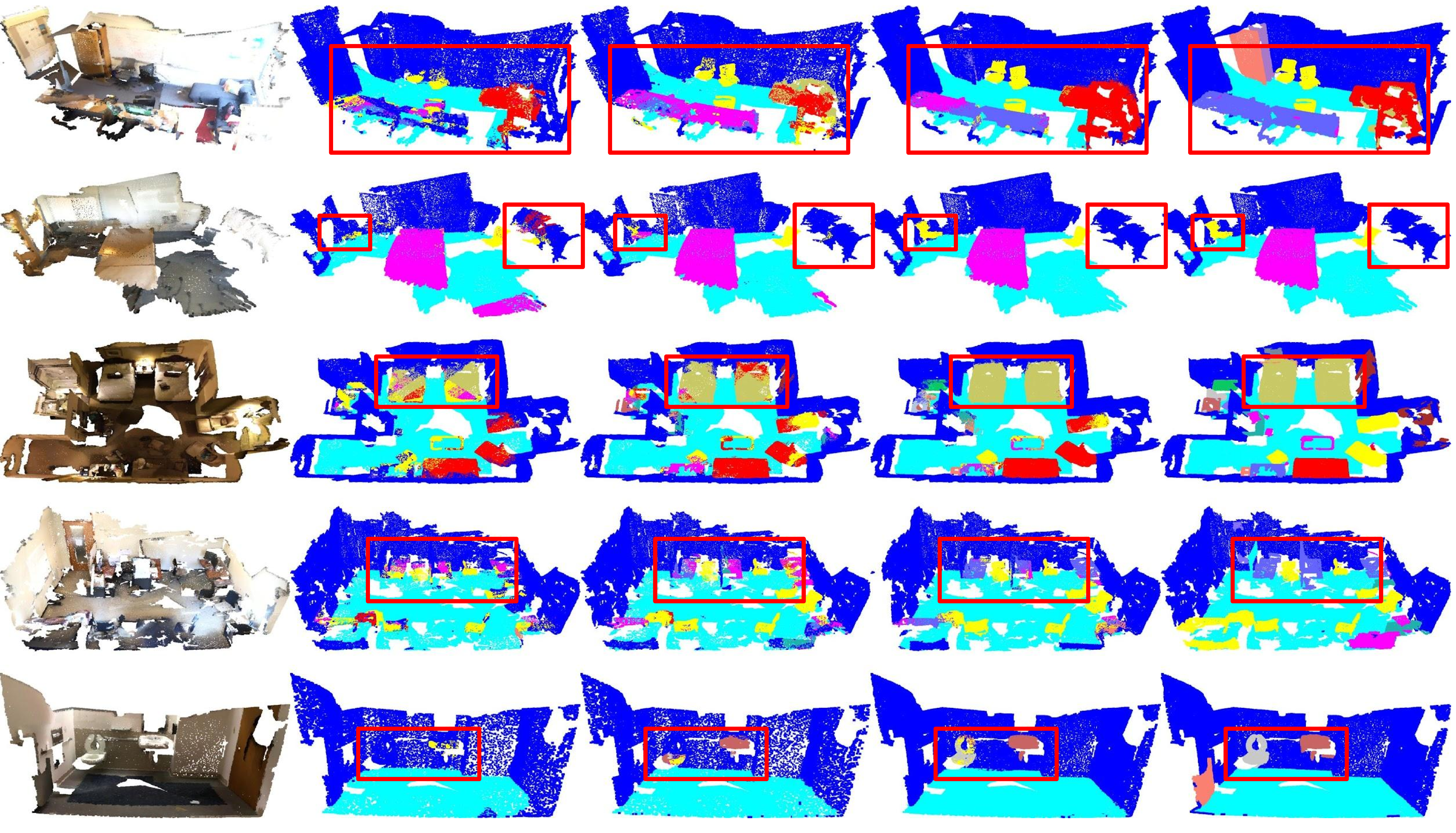}
\end{center}
   \caption{Sample segmentation results on the ScanNet dataset. From left to right are the input scenes, results produced by PointNet, PointNet++, RSNet, and ground truth. Interesting areas have been highlighted by red bounding boxes. Best viewed with zoom in.}
\label{fig:scannet}
\end{figure*}

In the ScanNet dataset, the previous state-of-the-art method is PointNet++ \cite{pointnet2}. It only uses the $xyz$ information of point clouds as inputs. To make a fair comparison, we also only use $xyz$ information in the RSNet. \cite{pointnet2} only reported the global accuracy on the ScanNet dataset. However, as shown in the supplementary, the ScanNet dataset is highly unbalanced. In order to get a better measurement, we still use mean IOU and mean accuracy as evaluation metrics as previous sections. We reproduced the performances of PointNet\cite{pointnet}  and Pointnet++\cite{pointnet2} (the single scale version) on the ScanNet dataset \footnote{ We reproduced the PonitNet and PointNet++ training on ScanNet by using the codes \href{https://github.com/charlesq34/pointnet}{here} and \href{https://github.com/charlesq34/pointnet2}{here} which are published by the authors. The global accuracy of our version of PointNet and PointNet++ are 73.69\% and 81.35\%, respectively.} and report them in Table.\ref{table:scannet} as well. As shown in Table.\ref{table:scannet}, the RSNet has also achieved state-of-the-art results on the ScanNet dataset. Compared with the PointNet++, the RSNet improves the mean IOU and mean accuracy by 5.09 and 4.60. Some comparisons between different methods are visualized in Fig.\ref{fig:scannet}. These visualizations show that as a benefit of the local dependency module, the RSNet is able to handle small details such the chairs, desks, and toilets in inputs.

\begin{table*}
\begin{center}
\setlength\tabcolsep{2.0pt}
\begin{tabular}{c|c|cccccccccccccccc}
\Xhline{3\arrayrulewidth}

Method & mean & aero & bag  & cap & car & chair & \begin{tabular}{@{}c@{}}ear \\ phone\end{tabular}   & guitar & knife & lamp & laptop & motor & mug & pistol & rocket & \begin{tabular}{@{}c@{}}skate \\ board\end{tabular}  & table\\
\Xhline{3\arrayrulewidth}
Yi \cite{shapenet} & 81.4 & 81.0 & 78.4 & 77.7 & 75.7 & 87.9 & 61.9 & 92.0 & 85.4 & 82.5 & 95.7 & 70.6 & 91.9 & 85.9 & 53.1 & 69.8 & 75.3\\
KD-net \cite{kd-net} & 82.3 &80.1 & 74.6 & 74.3 & 70.3 & 88.6 & 73.5 & 90.2 & \textbf{87.2} & 81.0 & 94.9 & 57.4 & 86.7 & 78.1 & 51.8 & 69.9 & 80.3 \\
PN \cite{pointnet} & 83.7 & \textbf{83.4} & 78.7 & 82.5 & 74.9 & 89.6 & 73.0 & 91.5 & 85.9 & 80.8 & 95.3 & 65.2 & 93.0 & 81.2 & 57.9 & 72.8 & 80.6\\
PN++ * \cite{pointnet2} & 84.6& 80.4 &80.9& 60.0& 76.8& 88.1& \textbf{83.7}& 90.2 &82.6& 76.9& 94.7& 68.0& 91.2& \textbf{82.1}& 59.9& 78.2& \textbf{87.5} \\
SSCNN \cite{spec} & 84.7 & 81.6 & 81.7 & 81.9 & 75.2 & 90.2 & 74.9 & \textbf{93.0} & 86.1 & \textbf{84.7} & \textbf{95.6} & 66.7 & 92.7 & 81.6 & \textbf{60.6} & \textbf{82.9} & 82.1\\
PN++ \cite{pointnet2} & \textbf{85.1} & 82.4 & 79.0 & \textbf{87.7} & 77.3 & \textbf{90.8} & 71.8 & 91.0 & 85.9 & 83.7 & 95.3 & \textbf{71.6} & \textbf{94.1} & 81.3 & 58.7 & 76.4 & 82.6\\
\hline
Ours  & \textbf{84.9} &  82.7& \textbf{86.4}&84.1&\textbf{78.2}&90.4&69.3&91.4&87.0&83.5&95.4&66.0&92.6&81.8&56.1&75.8&82.2 \\

\end{tabular}
\end{center}
\caption{Results on the ShapeNet dataset. PN++ * denotes the PointNet++ trained by us which does not use extra normal information as inputs.}
\label{table:shapenet}
\end{table*}

\begin{table*}[!ht]
\begin{center}
\setlength\tabcolsep{2.0pt}
\begin{tabular}{c|c|c|c|c|c|c}
\Xhline{3\arrayrulewidth}

 & \begin{tabular}{@{}c@{}}PointNet \\ (vanilla) \cite{pointnet}\end{tabular} & \begin{tabular}{@{}c@{}}PointNet \cite{pointnet} \\  \end{tabular} & \begin{tabular}{@{}c@{}}PointNet++ \\ (SSG) \cite{pointnet2}\end{tabular} & \begin{tabular}{@{}c@{}}PointNet++ \\ (MSG) \cite{pointnet2} \end{tabular} & \begin{tabular}{@{}c@{}}PointNet++ \\ (MRG) \cite{pointnet2} \end{tabular} & \begin{tabular}{@{}c@{}}RSNet \\  \end{tabular}    \\
\Xhline{3\arrayrulewidth}
Speed & 1.0 $\times$ & 2.2 $\times$ & 7.1 $\times$  & 14.1 $\times$ & 7.5 $\times$ & 4.5 $\times$ \\
Memory & 844 MB & - & - & - & - & 756 MB \\
\end{tabular}
.
\end{center}
\caption{Computation analysis between PointNet, PointNet++, and RSNet.}
\label{table:computation}
\end{table*}

\subsection{Segmentation on the ShapeNet Dataset}
In order to compare the RSNet with some other methods \cite{kd-net, shapenet, spec}, we also report the segmentation results of RSNets on the ShapeNet part segmentation dataset. The same RSNet as in Section 4.1 is used here. It only takes the $xyz$ information as the convention. Its performance are reported in Table.\ref{table:shapenet}. Table.\ref{table:shapenet} also presents the results of other state-of-the-art methods including PointNet \cite{pointnet}, PointNet++ \cite{pointnet2}, KD-net \cite{kd-net}, and spectral CNN \cite{spec}. The RSNet outperforms all other methods except the PointNet++\cite{pointnet2} which utilized extra normal information as inputs. However, the RSNet can also outperform PointNet++ when it only takes $xyz$ information. This validates the effectiveness of the RSNet.

\subsection{Computation Analysis}
We now demonstrate the efficiency of RSNets in terms of inference speed and GPU memory consumption. We follow the same time and space complexity measurement strategy as \cite{pointnet2}. We record the inference time and GPU memory consumption of a batch of 8 4096 points for vanilla PointNet and the RSNet using PyTorch on a K40 GPU. Since \cite{pointnet2} reported the inference speed in TensorFlow, we use the relative speed w.r.t vanilla PointNet to compare speeds with each other. The speed and memory measurements are reported in Table.\ref{table:computation}. 

Table.\ref{table:computation} show that the RSNet is much faster than PointNet++ variants. It is near 1.6 $\times$ faster than the single scale version of PointNet++ and 3.1 $\times$ faster than its multi-scale version. Moreover, the GPU memory consumption of the RSNet is even lower than vanilla PointNet. These prove that the RSNet is not only powerful but also efficient.

\section{Conclusion}
This paper introduces a powerful and efficient 3D segmentation framework, Recurrent Slice Network (RSNet). An RSNet is equipped with a lightweight local dependency modeling module which is a combination of a slice pooling, RNN layers, and a slice unpooling layer. Experimental results show that RSNet can surpass previous state-of-the-art methods on three widely used benchmarks while requiring less inference time and memory.

\appendix
\addcontentsline{toc}{section}{Appendices}
\section*{Appendices}

\begin{table*}
\begin{center}
\setlength\tabcolsep{2.0pt}
\begin{tabular}{c|c|c|cccccccccccccc}
\Xhline{3\arrayrulewidth}

Method & mIOU & mAcc & ceiling  & floor & wall & beam & column & window & door & chair & table & bookcase & sofa & board & clutter    \\
\Xhline{3\arrayrulewidth}
PointNet \cite{pointnet} & 47.71 & - & - & - & - & - & - & - & - & - & -  \\
Ours & \textbf{56.47} & 66.45 & 92.48 & 92.83 & 78.56 & 32.75 & 34.37 & 51.62 & 68.11 & 59.72 & 60.13 & 16.42 & 50.22 & 44.85 & 52.03 \\
\end{tabular}
.
\end{center}
\caption{6-fold validation results on the Large-Scale 3D Indoor Spaces Dataset (S3DIS). IOU of each category is also reported.}
\label{table:stanford_6}
\end{table*}

Here, we present more details of experimental settings and more experimental results and discussions. 

\section{More Results and Discussions on the S3DIS dataset}
In the S3DIS dataset, there are 272 indoor scenes captured from 6 areas in 3 buildings. The points are annotated in 13 categories. To process this dataset, our RSNet takes points with 9 dimensional features as inputs as in \cite{pointnet}. The first three, middle three, and last three dimensions represent the $xyz$ coordinates, RGB intensities, and normalized $xyz$ coordinates, respectively.

In the main text, we used the training/testing split in \cite{segcloud} is used to avoid dividing areas from same building to both training and testing sets. However, in \cite{pointnet}, the authors reported their performances using 6-fold validation. In order to comprehensively compare with \cite{pointnet}, we also present the 6-fold validation performances of RSNet in Table.\ref{table:stanford_6}. The results show that our RSNet outperforms PointNet by a large margin while requiring less memories and reasonable extra inference times.

Both Table.\ref{table:stanford_6} and the Table.1 in the main text show that while all the methods work well on some categories like ceiling, floor and wall, they all fail to achieve the same level of performances on the categories like beam, column, and bookcase. This is because the S3DIS dataset is a highly unbalanced dataset. From the data portion statistics in Table.\ref{table:stanford_stat} we notice that ceiling, floor and wall are the dominant classes which have 7 $\sim$ 50 times more training data than the rare classes. This makes the segmentation algorithms fail to generalize well on the rare classes. In order to alleviate this problem, we adopt the median frequency balancing strategy \cite{median} in our RSNet training. The results are compared with the baseline in Table.\ref{table:stanford_median}. It shows that using median frequency balancing improves performances in terms of the mean accuracy. However, there is a slight decrease in mean IOU.

\begin{table*}[ht]
\begin{center}
\setlength\tabcolsep{2.0pt}
\begin{tabular}{c|cccccccccccccc}
\Xhline{3\arrayrulewidth}

 & ceiling  & floor & wall & beam & column & window & door & chair & table & bookcase & sofa & board & clutter    \\
\Xhline{3\arrayrulewidth}

Data Per-centage (\%) & 25.3 & 23.3 & 17.3 & 2.42 & 1.6 & 1.1 & 4.6 & 3.4 & 5.3 & 0.5 & 3.3 & 0.7 & 11.2

\end{tabular}
.
\end{center}
\caption{Data portion of each category in the training set of the S3DIS dataset.}
\label{table:stanford_stat}
\end{table*}

\begin{table*}[ht!]
\begin{center}
\setlength\tabcolsep{2.0pt}
\begin{tabular}{c|ccccccccccc}
\Xhline{3\arrayrulewidth}

 & wall  & floor & chair & table & desk & bed & bookshelf   & sofa & sink & bathtub  \\
\Xhline{3\arrayrulewidth}
Data Per-centage (\%) & 36.8 & 24.9 & 4.6 & 2.5 & 1.7 & 2.6 & 2.0 & 2.6 & 0.3 & 0.3  \\

 & toilet & curtain & counter & door & window & shower-curtain  & refridgerator & picture & cabinet & other furniture \\
\Xhline{3\arrayrulewidth}
Data Per-centage (\%) & 0.3 & 1.5 & 0.6 & 2.3 & 0.9 & 0.2 & 0.4 & 0.4 & 2.6 & 2.5

\end{tabular}
.
\end{center}
\caption{Data portion of each category in the training set of the ScanNet dataset.}
\label{table:scannet_stat}
\end{table*}

\begin{table*}[ht!]
\begin{center}
\setlength\tabcolsep{2.0pt}

\begin{tabular}{cccccccccccc}
\Xhline{3\arrayrulewidth}

Method & mIOU & mAcc & wall  & floor & chair & table & desk & bed & \begin{tabular}{@{}c@{}}book- \\ shelf\end{tabular}   & sofa & sink\\
\Xhline{3\arrayrulewidth}
RSNet				& 39.35	& {48.37}	&{79.23}	& {94.10}	& \textbf{64.99} 	&\textbf{51.04} 	& {34.53}	&\textbf{55.95} &{53.02} &\textbf{55.41}& {34.84} \\
RSNet	with RGB   	& \textbf{41.16}	& \textbf{50.34}	&\textbf{79.38}	& \textbf{94.21}	& {63.65} 	&{48.67} 	& \textbf{35.27}	&{53.09} &\textbf{53.67} &{51.06}& \textbf{41.00} \\

\Xhline{3\arrayrulewidth}
Method  & bathtub & toilet & curtain & counter & door & window & \begin{tabular}{@{}c@{}}shower \\ curtain\end{tabular}  & \begin{tabular}{@{}c@{}}refrid- \\ gerator\end{tabular}  & picture & cabinet & \begin{tabular}{@{}c@{}}other \\ furniture\end{tabular}   \\
\Xhline{3\arrayrulewidth}
RSNet				& {49.38}	& {54.16}	&6.78	& \textbf{22.72}	& {3.00}	& {8.75}	&\textbf{29.92} 	& {37.90}	&{0.95}	&\textbf{31.29} &{18.98}\\
RSNet with RGB		& \textbf{60.37}	& \textbf{63.20}	&\textbf{8.30}	& {20.90}	& \textbf{15.32}	& \textbf{15.67}	&\textbf{24.36} 	& \textbf{39.76}	&\textbf{4.30}	&{30.06} &\textbf{20.98}\\
\hline
\end{tabular}

\end{center}
\caption{Results on the ScanNet dataset. IOU of each category is also reported here.}
\label{table:scannet_rgb}
\end{table*}

\begin{table}
\begin{center}
\begin{tabular}{c|c|c}
\Xhline{3\arrayrulewidth}

Method  & mIOU & mAcc    \\
\Xhline{3\arrayrulewidth}
 RSNet & \textbf{51.93} & 59.42\\
 RSNet-\emph{median} & 48.68  & \textbf{62.09} \\

\end{tabular}
.
\end{center}
\caption{Results of different training strategies on the S3DIS dataset.}
\label{table:stanford_median}
\end{table}



\section{More Results and Discussions on the ScanNet dataset}

The ScanNet dataset contain 1,513 scenes captured by the Matterport 3D sensor. We follow the official training/testing split \cite{scannet} in this paper. The points are annotated in 20 categories and one background class. As shown in Table.\ref{table:scannet_stat}, the ScanNet dataset is also highly unbalanced. Thus, we use the mean IOU and mean accuracy as evaluation metrics in the main text to better measure the performances for this dataset. To process the ScanNet dataset, out RSNet takes points with 3 dimensional features ($xyz$ coordinates) as inputs as in \cite{pointnet2}.

In order to further improve the performances on the ScanNet dataset, we train a RSNet taking not only $xyz$ coordinates but also RGB intensities as inputs. The results are reported in Table.\ref{table:scannet_rgb}. It shows that RGB information can slightly improve the performances of our baseline model. The mean IOU and mean accuracy are improved by 1.81 and 1.97. Moreover, detailed per-class IOUs show that the RGB information is particularly helpful for categories like door, window, and picture. These classes can be easily confused with walls when only geometric information ($xyz$ coordinate) is present. However, RGB information helps the network distinguish them from each other.

{\small
\bibliographystyle{ieee}
\bibliography{egbib}
}

\end{document}